# A Plant Root System Algorithm Based on Swarm Intelligence for One-dimensional Biomedical Signal Feature Engineering

Authors:

Rui Gong; Email: gong-rui@ed.tmu.ac.jp

Kazunori Hase; Email: kazunori.hase@tmu.ac.jp

**Abstract:**

To date, very few biomedical signals have transitioned from research applications to clinical applications. This is largely due to the lack of trust in the diagnostic ability of non-stationary signals. To reach the level of clinical diagnostic application, classification using high-quality signal features is necessary. While there has been considerable progress in machine learning in recent years, especially deep learning, progress has been quite limited in the field of feature engineering. This study proposes a feature extraction algorithm based on group intelligence which we call a Plant Root System (PRS) algorithm. Importantly, the correlation between features produced by this PRS algorithm and traditional features is low, and the accuracy of several widely-used classifiers was found to be substantially improved with the addition of PRS features. It is expected that more biomedical signals can be applied to clinical diagnosis using the proposed algorithm.

1. Introduction

One-dimensional signals are the most easily collected signals by machine. Such signals are widely used in industry, agriculture, finance, and, in particular, the medical field. Lately, massive one-dimensional biomedical signals have been digitized and stored as time-series data on hard disks, and much consideration has been given to how to make effective use of these data. Applying biomedical signals to clinical diagnosis is a frontier multidisciplinary research subject that has great significance for the advancement of medicine [1, 2]. Classification, which is the ability to group objects by their similar features, is one of the primary time-series data processing applications. Supervised classification, in particular, is considered to perform well, and it would seem that a robust and well-established classifier with supervised machine learning techniques can help doctors improve the accuracy of their clinical diagnoses [3, 4].

To develop a powerful classifier requires heavy training. If an untrained classifier can be regarded as an "infant," then features training can be regarded as the "experience" required for the infant to progress towards adulthood. Before classification, many correct and appropriate features need to be extracted and transformed from signals or images in a process referred to as feature engineering [5]. Feature engineering is a key to success in applied machine learning, as it allows feature sets to become more flexible and reduces overmatching or undermatching in classification models, thus improving classification accuracy [5]. The features of one-dimensional time-series signals, especially biomedical signals, can be roughly divided into time-domain features, frequency-domain features, and nonlinear features [6]. These features appear to be unrelated; however, features derived from operations and transformations of the original signal can be restored to the original signal by inverse operations and inverse transformations. Even when these features are carefully selected, independence is still not satisfied, bringing bias to the training process. To counteract this anchoring bias, we propose a plant root system (PRS) algorithm based on swarm intelligence as an auxiliary feature.

Franz et al. [7, 8] hypothesized that the root system of plants is a swarm intelligence system and that such systems are not limited to animals and humans. According to this hypothesis, root apices communicate with each other through electrical signals or chemical pheromones, and then build a massive "brain-like" system in the underground world [9]. Within the complex root system, independent roots acquire water and nutrients through cooperation. At the same time, to support proper growth and development, the root system smartly competes with other root systems and chooses to start a "war" when it has an advantage or retreat when it is at a disadvantage [7, 10]. In this article, the proposed PRS approach mimics the growth of plant roots, where the root apices cooperate to absorb as much of the sustaining nutrients as possible.

In our proposed approach, the base features used by the selected machine-learning model to classify biomedical signals are equivalent to nutrients in the digital soil. These base features are features obtained from common feature extraction methods. In addition to these base features, the sum of the nutrients absorbed during growth serves as a first auxiliary feature; the growing area of the root system is the second auxiliary feature. These auxiliary features are referred to as PRS auxiliary features. In the model training process, the most significant advantage of the PRS features based on swarm intelligence is that they are not only ideally independent (i.e., they have low correlation with the other features), but they are also highly correlated with the dependent variable. The independence of the PRS features derives from the brain-like structure of the root system, where the co-determination of root tips reduces the correlation with the base features. When these PRS auxiliary features are appended to the training set, the increase in independence between features results in a decrease in bias and variance and an increase in the accuracy of classification by machine learning. For biomedical signals, an increase in classification accuracy means an increase in the positive diagnosis rate, which ultimately leads to a greater use of more biomedical signals in the clinical field.

2. Method

2.1 Extraction of base features

The growth of natural plants requires suitable soil; similarly, extraction of PRS features requires digital soil composed of base features. When extracting base features, time-domain features are generally preferred to frequency-domain features and nonlinear features. For one-dimensional biomedical signals with a non-stationary nature, the time-domain feature involves a lighter computational burden and has stronger reliability than nonlinear features [11]. Moreover, time-domain features also avoid the risk of spectral leakage caused by signal decomposition and is more dependable than frequency-domain features [12].

Time-domain feature selection is an extremely important process for base feature extraction. The number of features needs to be carefully considered: too few features will limit the growth of the root system, while too many features may adversely affect the performance of the classifier [13]. To comprise the set of base features, we selected 12 features that have relatively low computational complexity and that have been successfully used for classification with a high degree of accuracy [14–16]. Definitions of the base features are given in Table 1, where $x_n$ represents the one-dimensional biomedical signal in segment $n$, N is the length of signal, and $\bar{x}$ is the average value of $x_n$.

Table 1 Time-domain features list

| First-level features | Definition | Equation |
|---|---|---|
| Standard deviation (STD) | $\text{STD} = \sqrt{\frac{1}{N}\sum_{i=1}^{N}\left\lvert x_i - \frac{1}{N}\sum_{i=1}^{N} x_i \right\rvert^2}$ | (1) |
| Variance of signal (VAR) | $\text{VAR} = \frac{1}{N-1}\sum_{n=1}^{N} x_n^2$ | (2) |
| Root mean square (RMS) | $\text{RMS} = \sqrt{\frac{1}{N}\sum_{n=1}^{N} x_n^2}$ | (3) |
| Skewness (SKW) | $\text{SKW} = \frac{1}{N}\sum_{n=1}^{N}(x_n-\bar{x})^3 / \left(\frac{1}{N}\sum_{n=1}^{N}(x_n-\bar{x})^2\right)^{3/2}$ | (4) |
| Kurtosis (KURT) | $\text{KURT} = \frac{1}{N}\sum_{n=1}^{N}(x_n-\bar{x})^4 / \left(\frac{1}{N}\sum_{n=1}^{N}(x_n-\bar{x})^2\right)^2$ | (5) |
| Mean absolute value (MAV) | $$\mathrm{MAV} = \frac{1}{N}\sum_{n=1}^{N} |x_n|$$ | (6) |
| Zero crossing (ZC) | $$\mathrm{ZC} = \sum_{n=1}^{N-1} [\mathrm{sgn}(x_n \times x_{n+1}) \cap |x_n - x_{n+1}| \geq \text{threshold}]$$ | (7) |
| Slope sign change (SSC) | $$\mathrm{SSC} = \sum_{n=2}^{N-1} [f[(x_n - x_{n-1}) \times (x_n - x_{n+1})]]$$ $$f(x) = \begin{cases} 1, & \text{if } x \geq \text{threshold} \\ 0, & \text{otherwise} \end{cases}$$ | (8) |
| Willison amplitude (WAMP) | $$\mathrm{WAMP} = \sum_{n=1}^{N-1} f(|x_n - x_{n+1}|)$$ $$f(x) = \begin{cases} 1, & \text{if } x \geq \text{threshold} \\ 0, & \text{otherwise} \end{cases}$$ | (9) |
| Simple Sign Integral (SSI) | $$\mathrm{SSI} = \sum_{n=1}^{N} |x_n|^2$$ | (10) |
| Non-linear energy (NLE) | $$NLE(x_n) = \frac{1}{N-2}\sum_{i=2}^{N-1} x_n(i)^2 - x_n(i-1)x_n(i+1)$$ | (11) |
| Waveform length (WL) | $$\mathrm{WL} = \sum_{n=1}^{N-1} |x_{n+1} - x_n|$$ | (12) |

Among the 12 features comprising the set of base features, the most basic are the statistical features. These include variance (VAR) and standard deviation (STD), the most common statistics; the root mean square (RMS) amplitude, a statistical measure of the magnitude of a time-varying quantity; skewness (SKW), a measure of the asymmetry of the probability distribution of a real-valued random variable; kurtosis (KURT), the scaled form of the fourth moment and a measure of "peakedness"; and mean absolute value (MAV), which is similar to the average rectified value.

Three features related to frequency are also included: zero crossing (ZC), when the amplitude value of a signal crosses the zero amplitude axis; slope sign change (SSC), representing the frequency information of a signal in the time domain; and Willison amplitude (WAMP), the sum of the difference between the signal amplitude for two adjacent samples.

The final three features are indicators of energy and complexity: The simple sign integral (SSI) is determined as the energy of a signal segment; nonlinear energy (NLE) is the approximative energy of signal amplitude; and waveform length (WL) serves as a measure of signal complexity.

2.2 Features sorting

For the development of a root system, the site where the seed is planted needs nutritious soil. In this study, the distribution of nutrients is equivalent to the distribution of the base features.

Each element $\beta_{m,n}$ of the base features set $B_{m\times n}\langle\;\rangle$ needs to be sorted. Before this is done, the dataset must be normalized using a scaling technique in the pre-processing stage to eliminate scale differences among the various elements when sorting. In this study, min-max normalization was to scale the data in the range $[0,1]$. The advantage of this method is that it can maintain the relationships that exist among the original data [17]. Here, the min-max normalization based on the features set $\dot{B}_{m\times n}\langle\;\rangle$ can be shown as

$$\dot{B}_{m\times n}\langle\;\rangle = \frac{\beta_{m,n} - B\langle\;\rangle_{min}}{B\langle\;\rangle_{max} - B\langle\;\rangle_{min}} \tag{13}$$

where $n = 1, 2, \cdots, 12$ and $m = 1, 2, \cdots, length\ of\ \beta_n$.

Feature importance measurement, or feature selection, provides the means by which the initial number of features can be reduced by eliminating those features with a low importance score for improving the performance of the classifier [18]. In this study, we chose not to delete features, but rather to sort features based on their worth score. To establish a feature's worth score, we measured the information gain associated with the feature; the features with higher worth scores were those closer to the center location. Information gain was first shown in a decision tree and used to rank the priority of feature nodes; this was then expanded and applied to our importance measurement [19].

Let the class label of a feature have two distinct values defining binary classes $\{C_1, C_2\}$. $Info\ (\dot{B})$, also known as the entropy of $\dot{B}_{m\times n}\langle\;\rangle$, can be defined as follows:

$$Info\ (\dot{B}) = -p_1 log_2\ (p_1) - p_2 log_2\ (p_2) \tag{14}$$

where $p_1, p_2$ is the nonzero probability that an arbitrary tuple in $\dot{B}_{m\times n}\langle\;\rangle$ belongs to class $C_1, C_2$ and is estimated by $|C_{1\ or\ 2,\dot{B}}|/|\dot{B}|$.

Since each feature column $\dot{\beta}_n$ is a set of discontinuous values, these discrete values of $\dot{\beta}_n$ need to be split into two sets using an unsupervised algorithm. The K-means clustering method is a good choice for partitioning the given data set into k groups, where k represents the two groups pre-specified by the analyst [20]. In this study, this process is repeated 25 times to produce lower within-cluster variation and a more stable result. Each $\dot{\beta}_{m,n}$ now belongs to splitting set $\{s_n^1, s_n^2\}$ and has a class label. Then the expected information required to classify the tuple from $\dot{B}$ based on feature $\dot{\beta}_n$ is

$$Info_{\dot{\beta}_n}(\dot{B}) = \sum_{i=1}^{2} \frac{|\dot{B}_{s_n^i}|}{|\dot{B}|} \times Info(\dot{B}_{s_n^i}) \tag{15}$$

$$Gain\ (\dot{\beta}_n) = Info\ (\dot{B}) - Info_{\dot{\beta}_n}(\dot{B}) \tag{16}$$

The $\dot{\beta}_n$ with the highest information gain is chosen to replace the center of $\{\dot{\beta}_1, \dot{\beta}_2, \cdots, \dot{\beta}_{12}\}$, and the other features are sorted, one-by-one, from the center to the two sides, with decreasing information gain. In this way, a sorted base feature set is obtained.

2.3 Generating nutritious soil based on features.

In this study, the digital soil consisting of the base features set cannot have only width and not depth. Therefore, continuous feature discretization is necessary (from scalar to vector). In this process, equal width interval binning is perhaps the most common method to discretize data for producing nominal values from continuous features [21]. With this binning method, if a feature $a$ is observed to have values bounded by $a_{max}$ and $a_{min}$, then the method computes $k$ equally-sized bin widths

$$\delta = \frac{a_{max} - a_{min}}{k} \tag{17}$$

and constructs bin boundaries at $a_{min} + j\delta$ where $j = 1, 2 \cdots, k-1$. In order to be closer to the natural environment of the root system, we set $k = 15$.

In addition, the vertical distribution of soil nutrients should follow, to the extent possible, the laws of nature. Most nutrients are concentrated in the shallowest layer and decrease with depth since nutrients return to the soil through biomedical cycles [22]. Thus, the discrete feature set $d_{15 \times 12}$ is arranged in decreasing order from top to bottom. On the other hand, the horizontal distribution of nutrients is associated with the crustal movement. The kernel convolution values $\tau_A$ and $\tau_B$ are used to calculate the reconstituted nutrient distributions.

The nutritional reconstructions are divided into two types: $\tau_A$ affects the nutrient distribution from shallower layers, while $\tau_B$ affects the nutrient distribution from deeper layers. The convolution kernel is then be defined as

$$\tau_A = \frac{1}{2}\begin{bmatrix} 1 & 1 & 1 \\ 0 & 1 & 0 \\ 0.5 & 0.5 & 0.5 \end{bmatrix} \tag{18}$$

$$\tau_B = \frac{1}{2}\begin{bmatrix} 0.5 & 0.5 & 0.5 \\ 0 & 1 & 0 \\ 1 & 1 & 1 \end{bmatrix} \tag{19}$$

The nutritious soil used to generate the root system is written as $D_{15\times12}$, which results from the convolution of $d_{15\times12}$ with kernel $\tau_A$ and $\tau_B$. The calculated target matrix requires zero-padding with size one before each convolution.

The completed soil with the mineral nutrition generation process is shown in Fig. 1. Hereinafter, the matrix for the distribution of nutrients in the soil will be referred to as the "nutrient matrix."

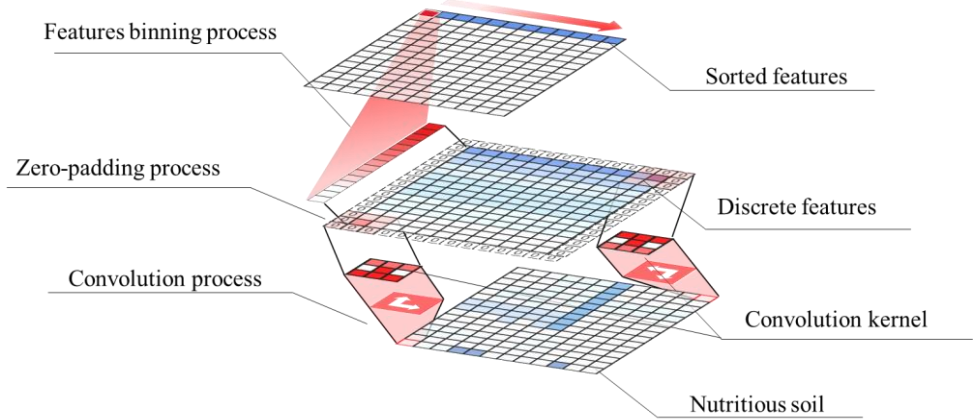

Fig. 1 Top-to-bottom order of soil matrix construction. The elements of the base feature set are arranged in order of importance, from the center to either side. Each element in the set is then discretized. Finally, the soil matrix is obtained from two convolution calculations.

2.4 Plant root system algorithm

In plants, the radicle, or primary root, is first organ to emerge from the seed coat. Before the first leaf grows, the energy needed by the cotyledon for root development all comes from the seed itself. Additional inorganic nutrition (including water) is also essential; only a small part of this nutrition comes from the seed itself, with the remainder coming from the surrounding soil containing mineral nutrients [23, 24]. The rule inherent in the genes of higher plants is to develop a sufficiently large first leaf and sufficiently long roots before the energy stored in the seeds is exhausted. The algorithm we propose follows this same rule. Since organic matter cannot be synthesized through photosynthesis before the first leaf has grown, the task is to absorb more mineral nutrients by growing a sufficient root system using limited energy (root division) in a limited time. Hereinafter, the matrix distribution for the roots in the soil will be referred to as the "root matrix."

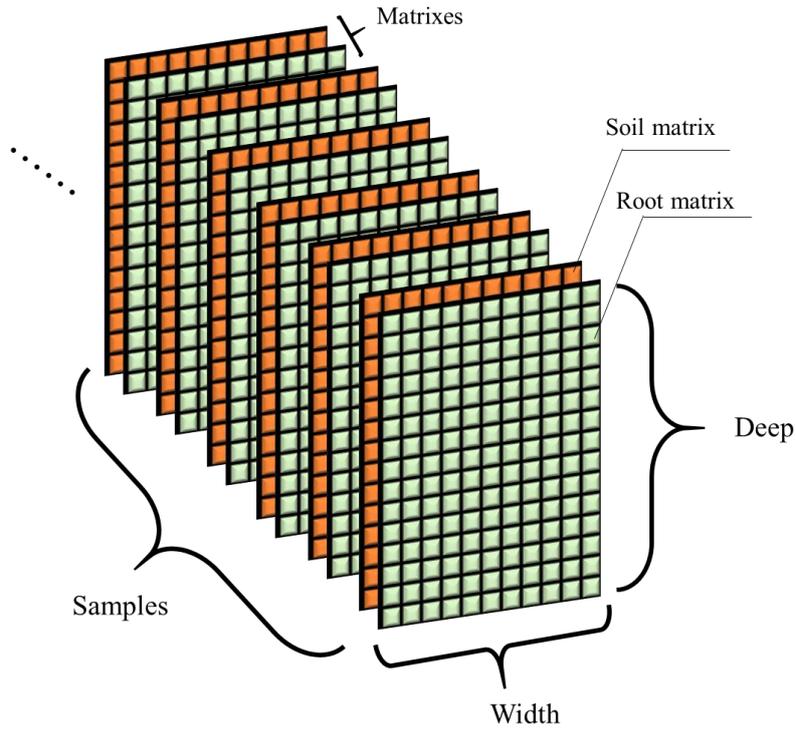

Fig. 2 A 4D data tensor of biomedical signals

In our proposed approach, the above natural processes need to be transformed into a data processing program. First, the root matrix and the nutrient matrix form a 4-dimensional tensor, as shown in Fig 2. The initial root matrix is a zero matrix, where the upper center location assigning values can create a radicle matrix. To absorb more nutrients, the rule of root division is to preferentially proliferate at the location of the global maximum in the mapping nutrient matrix; at the same time, location neighbor root-tips are necessary. The radicle matrix, photosynthetic day, and daily root division are then available as parameters to be modified. The development process can be characterized by the root distribution area and total nutrient absorption. Fig. 3 shows the natural process and digital process of building a brain-like root system. In Fig. 3, we can see that the growth of digital roots is dependent on the base features. Usefully, by drawing out PRS features from the brain-like root system, correlation with the base features is reduced.

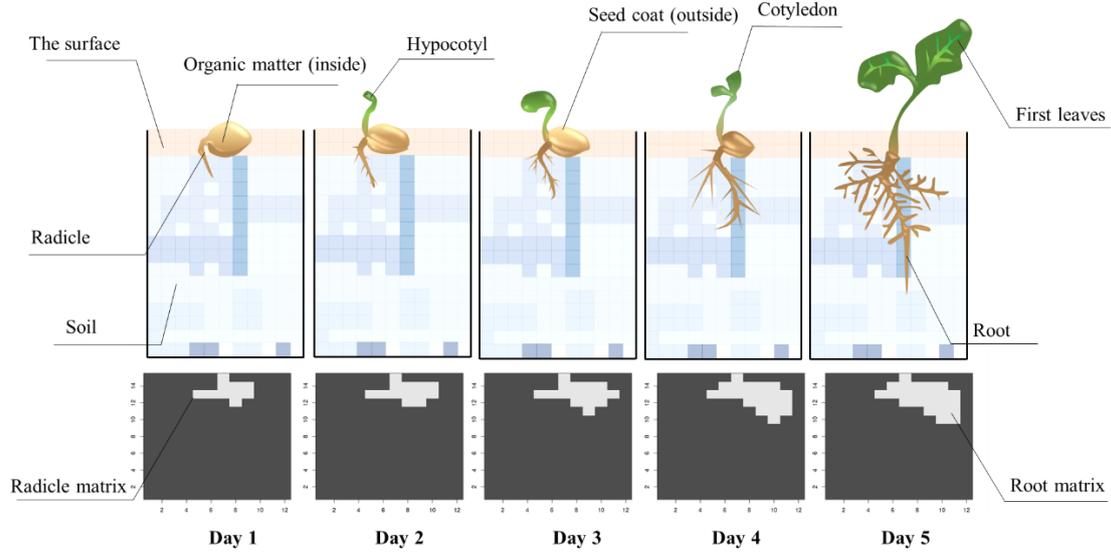

Fig. 3 The natural process and digital process of building brain-like root systems.

The growth process of the roots can be written as pseudocode, as shown in Table 2. In the proposed algorithm, the core steps of root development involve seeking the global maximum location of the nutrient matrix map relative to the root matrix location using a tensor of the base features set of a biomedical signal. The entire root system makes its decisions jointly in brain-like fashion to improve total nutrient absorption.

Table 2 Plant root system algorithm

| **Algorithm 1** Plant root system | |
|---|---|
| 1  Input: | ▷ x, y are the number of rows and |
| 2    $\mathbb{DR}_{x \times y}$: distribution of the radicle in the soil (initial root matrix); | columns of the distribution matrix |
| 3    $\mathbb{DN}_{x \times y}$: distribution of nutrients in the soil (nutrient matrix); | respectively |
| 4  Output: | |
| 5    $na$: nutrients absorbed by the roots; | |
| 6    $A$: root distribution area; | |
| 7  $na \leftarrow 0$; | ▷ $\mathbb{R}_{x \times y}$: distribution of the growing |
| 8  $\mathbb{R}_{x \times y} \leftarrow \mathbb{DR}_{x \times y}$; | root in the soil (root matrix) |
| 9  **for** $i = 1 \to day$ **do** | ▷ day is the days for first leaf to appear |
| 10   **for each** $\mathbb{R}_{[x^{th}, y^{th}]} = 1$ **do** | ▷ $x^{th} \in (1,2, \cdots x) \; y^{th} \in (1,2 \cdots y)$ |
| 11     $m \leftarrow \{\mathbb{DN}_{[x+1^{th}, y^{th}]}, \mathbb{DN}_{[x-1^{th}, y^{th}]}, \mathbb{DN}_{[x^{th}, y+1^{th}]}, \mathbb{DN}_{[x^{th}, y^{th}-1]}\}$; | ▷ $\mathbb{R}_{[x^{th}, y^{th}]} \in \mathbb{R}_{x \times y} \; \mathbb{DN}_{[x^{th}, y^{th}]} \in \mathbb{DN}_{x \times y}$ |
| 12     **return** $m$ to $\mathbb{M}$; | ▷ $\mathbb{M}$ is a matrix |
| 13   **end for each** | |
| 14   **decreasing sort** $\mathbb{M}$ to $M$; | ▷ $M$ is a vector |
| 15   **for** $j = 1 \to l$ **do**; ($l \leq length\ of\ M$) | ▷ $l$ is a limit of root division in one day |
| 16    $M[j]$ **mapping back to** $[x^{th}, y^{th}]$; | |
| 17    **if** $\mathbb{R}_{[x^{th}, y^{th}]} \neq 1$ **do** | |
| 18      $\mathbb{R}_{[x^{th}, y^{th}]} \leftarrow 1$; | |
| 19-21   $na \leftarrow \begin{cases} 0, & M[j] = 0 \\ \frac{M[j]}{1+|M[j]|} + 0.49, & M[j] \neq 0 \end{cases}$ | ▷ absorption rate |
| 22    **end if** | |
| 23    **do** $na++$; | |

```
24      return na and ℝ_{x×y};
25    end for
26    do ℝ̇ ← ℝ;                              ▷ the root extraction as polygon
27    do A ← area (ℝ̇);                       ▷ area of the polygon
28    return A;
29  end procedure
```

In this procedure, the roots distribution can be drawn in a coordinate system as a polygon. The area of this polygon serves as a feature. For the polygon area calculation, we used the tool developed by Bourke [25], which is one of the most well-known of such tools. The procedure outputs two results: the total amount of nutrients absorbed during root system development, designated the nutrition feature (NF), and the area of the roots distribution, designated the root feature (RF). Both are referred to as PRS features.

$$A = \frac{1}{2} \sum_{i=0}^{N-1} (p_i q_{i+1} - p_{i+1} q_i) \tag{20}$$

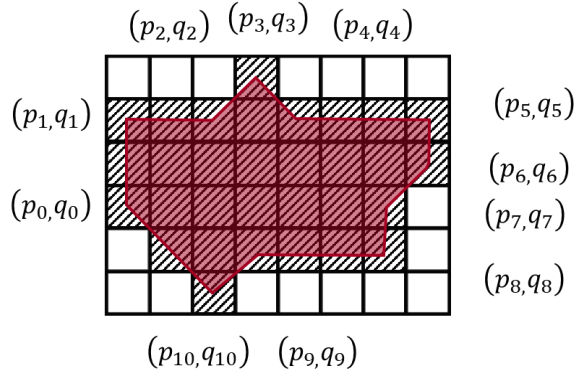

Fig. 4 An example of a 10-vertex closed polygon constructed from a root matrix

The calculation of NF and RF is shown as *na* (nutrient amount) and *A* (area), respectively, in Table 2. The calculation of A is derived from Equation 20 and Fig. 4, where the polygon is closed and made up of line segments between N vertices $(p_i, q_i), i = 0$ to $N = 1$.

Incorporating the extracted PRS features into the unprocessed base features set produces the set for classification, composed of 14 features. Lastly, the final PRS set is obtained by again performing a min-max normalization.

3. Datasets and evaluation methods

In this study, the accuracy of various classifiers is used for evaluation. The key measurable for each classifier is the change in accuracy before and after adding the root-based features to the original feature set. The formula used for quantifying this binary accuracy is

$$\text{Accuracy} = \frac{TP + TN}{TP + FP + TN + FN} \tag{21}$$

where TP = True positive; FP = False positive; TN = True negative; FN = False negative.

3.1 Datasets

When selecting a dataset of biomedical signals, a non-stationary signal is preferred. In fact, nearly all biomedical signals are non-stationary since the human system is always in dynamic equilibrium under the regulation of the brain [26]. Accordingly, three biomedical signals were selected in our study: the vibroarthrographic signal (VAG), which is the vibration signal generated by the friction between the knee cartilage during flexion-extension movement to detect knee joint disorder [27, 28]; the electromyography signal (EMG) of the forearm, which is the electrical signal generated by the extensor carpi radialis longus during the fist-relax movement [29]; and the audio signal collected via a smartphone web app [30] of a cough coming from lungs and airways that may or may not be affected by COVID-19. All the above signals are binary class signals—young or old for the VAG signals, rest or fist for the EMG signals, and positive or negative for the cough sound signals. Table 3 describes the properties of these signals in detail.

Table 3 properties of the biomedical signals

| Dataset | Signal type | Class | Sample size | Sampling rate | Length | Noise reduction |
|---|---|---|---|---|---|---|
| VAG of knee joint | Vibration signal | 2 | 144 | 2000 Hz | 3k–5k | Yes |
| EMG of hand | Electrical signal | 2 | 72 | 1000 Hz | 3k–5k | No |
| Sound of coughing | Sound signal | 2 | 121 | 22.5 kHz | 60k–80k | No |

3.2 Evaluating the PRS features

    3.2.1 Verification by different classifiers

Generally speaking, the selection of classifiers for feature testing should follow the principle of interpretability, excluding black boxes to the extent possible [31, 32]. As the principal linear classifier in this study, we chose the logistic regression (LR) model. LR has been widely used in classification problems and offers the benefit of outputting probabilities. However, an LR classifier is also limited by these probabilities. A result of 51% or even 99% pointing to a given class can produce mismatching [31]. In contrast to LR, which focuses on maximizing the probability of group membership, a support vector machine (SVM) seeks to find the separating hyperplane that maximizes the distance of the closest points to the margin. Since the feature sets involve more than 12 dimensions, we chose a polynomial kernel-based nonlinear SVM as the main nonlinear classifier in the study. In addition, we used linear discriminant analysis (LDA) and quadratic discriminant analysis (QDA) as complementary classifiers to the LR and SVM classifiers. Together, these four machine learning-based classifiers were used to test the feature set.

### 3.2.1 Extraction method for the comparison set features

Feature extraction methods for biomedical signals commonly use both time-domain features and frequency-domain features. Since all the base features in this study are time-domain features, we used two frequency-domain features to complete our comparison feature set [6]. Specifically, we used max power spectrum density (MaxPSD) and median power spectrum density (MedPSD) as the auxiliary features to the base features in the comparison set. The two frequency-feature extraction techniques were based on the Fourier transform [6]. By incorporating the extracted MaxPSD and MedPSD measures into the unprocessed base features set, the comparison set included the same number of features (14) as the experimental set. As a final step, the comparison set was normalized using min-max normalization. Having established the comparison set, we then proceeded to compare the accuracy of the PRS sets and the comparison set.

Implementation of the signal processing, feature engineering, the PRS algorithm, and the classifiers was supported by RStudio and related packages [33–36].

## 4. Results

Four datasets were input into each of the classification models chosen for the study: The first is the base set containing only the 12 time-domain features. The second (base + NF) and third (base + RF) sets each contain one PRS feature and the 12 time-domain features, for a total of 13 features. The final set is the PRS set with 14 features.

All four datasets from the VAG signal, the EMG signal, and the cough signal were input, one by one, into the LR classifier, the SVM classifier, the LDA classifier, and the QDA classifier. Each classifier was trained with learning rates of 40% to 80% of the full samples (in increments of 10%). The remaining samples were used as test data. Classification accuracy was then calculated using Equation 21. The results for the VAG signal, EMG signal, and cough signal are summarized in Figs. 5 - 7. The vertical coordinate in the figures is the average accuracy of 100 classifications using the various learning rate with one-way ANOVA test after Shapiro test.

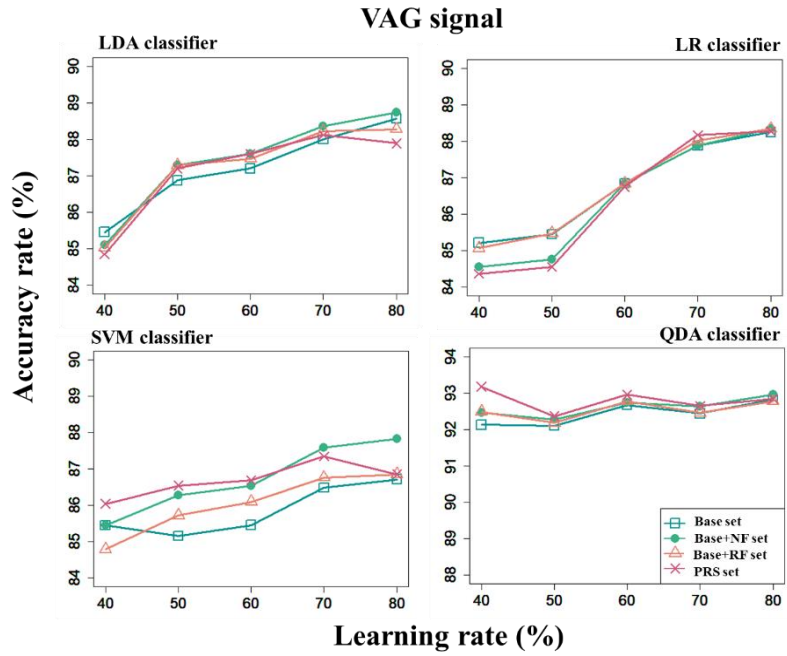

Fig. 5 Accuracy of the VAG signal for the four classifiers (significant, $F \in [4.22, 4.58]$, $p \leqq 0.0455$)

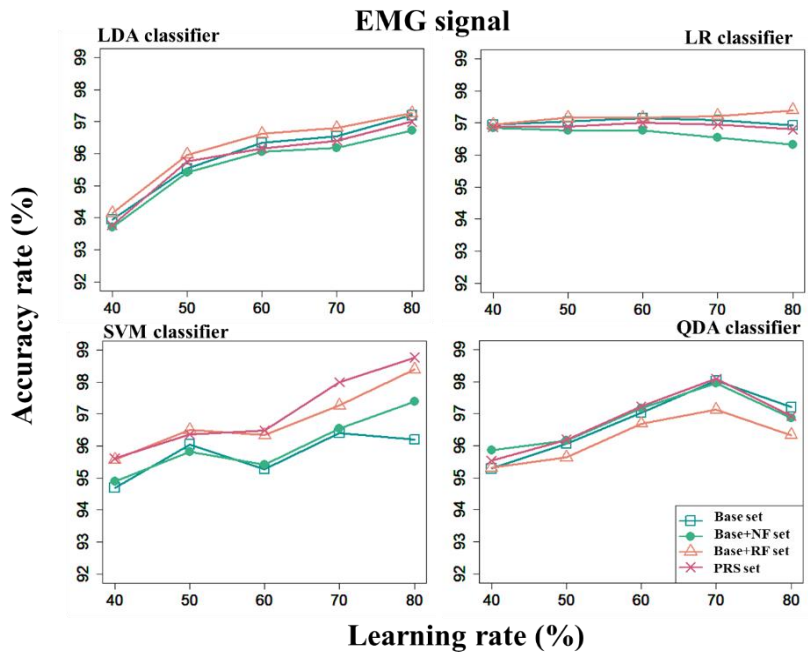

Fig. 6 Accuracy of the EMG signal for the four classifiers (50% LR classifier: quasi-significant, $F = 2.296$, $p = 0.0722$; Others: significant, $F \in [4.04, 4.39]$, $p \leqq 0.05$)

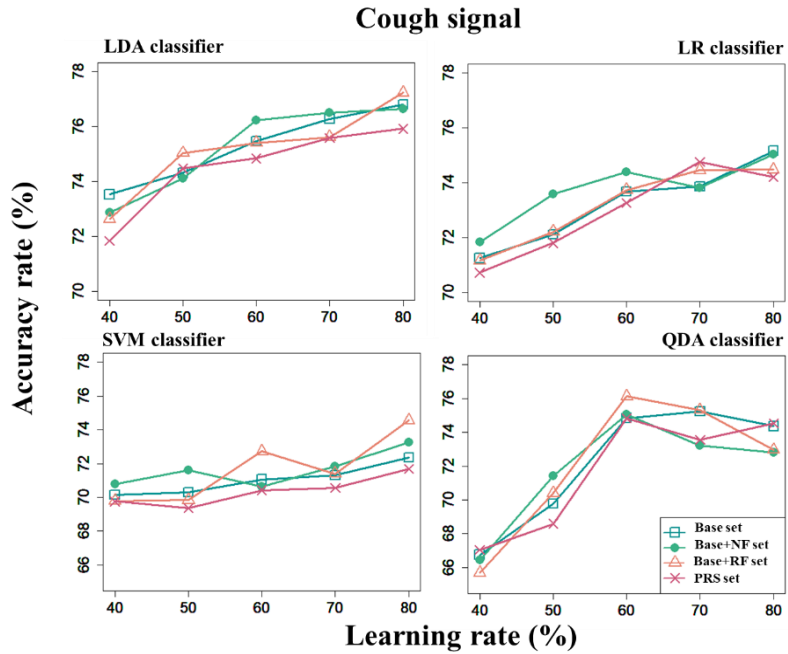

Fig. 7 Accuracy of the Cough signal for the four classifiers (significant, $F \in [4.24, 4.79]$, $p \leqq 0.05$)

As can be seen in the above figures, all four feature sets for the three types of biomedical signals show an increasing trend in the accuracy rate when the training rate is increased. However, what stands out most in the results is that the sets that include PRS features show a greater increase in accuracy than when only the base set is used. In addition, the accuracy for the base set is generally lower than the accuracy of at least one of the sets that include PRS features when the training rate is higher than 50%.

As shown in Fig. 5, for the VAG signal, the sets containing the nutritional feature (NF) tend to perform better. Here, the QDA classifier with the set containing both the nutritional feature and the root feature has the highest classification accuracy. Further analysis indicates that when the training rate is increased to 80% for the LR classifier, classification accuracy using the base set improves by 3.12%, while that for the base+NF set improves by 3.65%. Fig. 6 indicates that the classification accuracy for the EMG signal is particularly high for all four classifiers, in some cases as high as 98%, a value much higher than the maximum values for the other two signals. Somewhat surprisingly, while most sets containing the RF achieved higher accuracy than the base set, the base set with an 80% learning rate led the QDA classifications. Fig. 7 shows that the cough signal is less accurate and regular than the other signals. Here, the set containing the NF and the set containing the RF show higher accuracy, alternating with different training rates. In addition, the number and intensity of monotonicity changes are higher than those of the other signals.

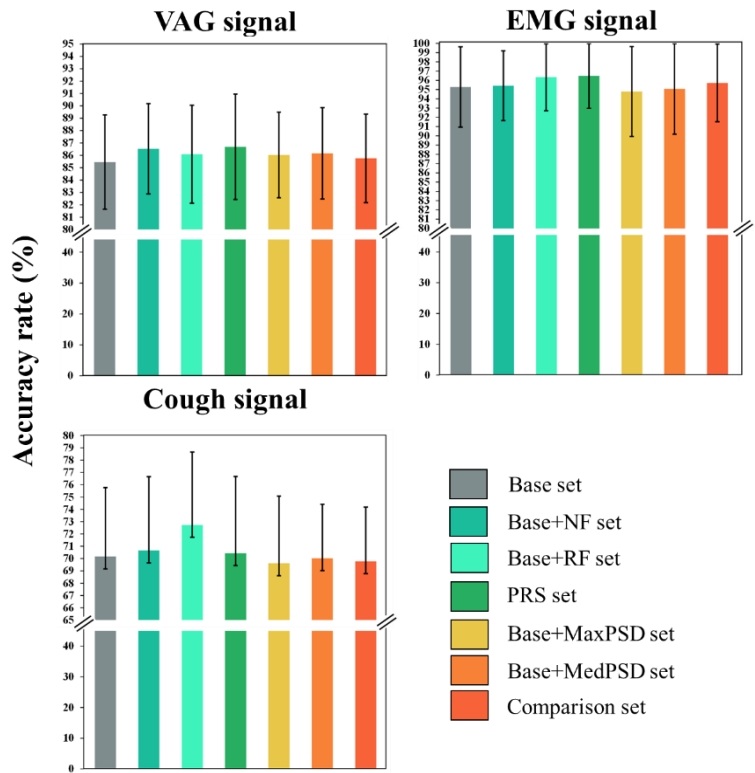

Fig. 8 The accuracy of the PRS sets and comparison sets with a 60% learning rate for the SVM classifier. The vertical coordinate is the average accuracy for 100 classifications; the error bar indicates the standard deviation. (p≦0.05)

Fig. 8 compares the accuracy of the base set, the sets containing the PRS features, and the sets containing frequency domain features using a 60% learning rate for the SVM classifier. A Tukey multiple comparisons test of means using a 95% family-wise confidence level was performed on the raw data shown in the figure. For the VAG signal, the average differences between the sets containing PRS features and the base set were all greater than 3.71%; the average differences for the sets containing PRS features and the sets containing the frequency domain features were all greater than 1.63%. All the p-values were less than 0.05. The EMG signal and the cough signal followed this same pattern.

Overall, our results indicate that the sets containing only the base features had a consistently lower accuracy rate and that the sets containing the PRS features had a consistently higher accuracy rate.

5. Discussion

For VAG signals, the sets containing PRS features showed a maximum accuracy of 93%, which is a higher rate than has been reported in all relevant past studies [27, 28]. These results agree with Rauber's (2015) findings, which showed that using only appropriate features could improve the performance of classifiers [13]. For cough signals, although the sets containing PRS features did not achieve the 80% accuracy reported by Chaudhari (2015), who used thousands of samples and deep learning to produce his results, our method achieved 76% accuracy with only one-tenth of the sample size [30]. We were unable to find published classification results for EMG signals; however, we believe that the classification accuracy we achieved of up to 98% is quite satisfactory.

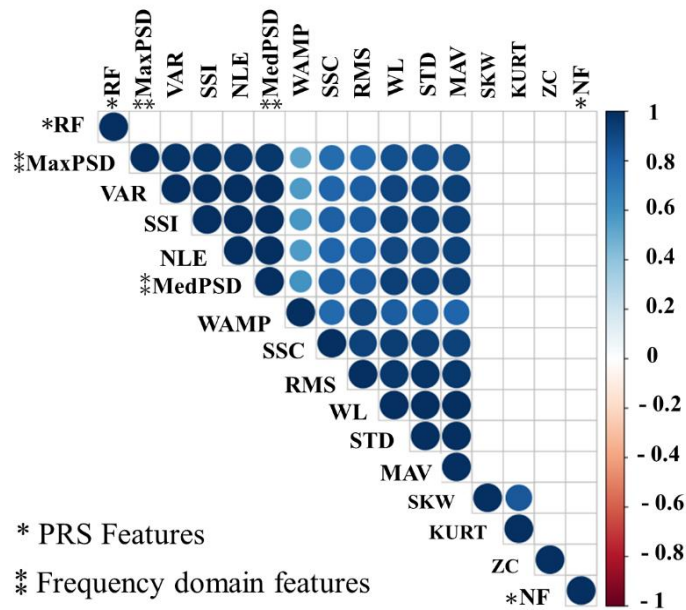

Fig. 9 Correlation between individual features of the EMS signal. The circle at the intersection of the rows of a feature and the columns of other features represents the correlation between the two features. Positive correlation has a cool tone circle and negative correlation has a warm tone circle.

The most significant characteristic of this study is its creative use of a swarm intelligence approach to draw out a feature set with exceptionally low correlations by simulating the development of natural plant root systems. The features correlation for the EMS signals is shown in Fig. 9. Surprisingly, two features based on the PRS algorithm have almost zero correlation with the other features. Conversely, the time-domain features and frequency-domain features are more or less correlated with the other features. The low correlation between the PRS features and the other features is due to the different feature extraction methods. The development of the root system is a product of group decision making in a process that reduces the correlation of the root matrix and the nutrient matrix in a tensor. It is for this reason that the performance of the classifier improves after adding the PRS features.

Another significant advantage of the PRS algorithm is its simplicity and interpretability. Anyone who has seen the root systems of plants in nature can understand and visualize the process of root development. Although the full procedure of the proposed method might appear to be rather lengthy, the algorithm is actually rather lightweight. The construction of the nutrient matrix comprises 2/3 of the work, while the construction of the root matrix and the extraction of PRS features make up the other 1/3. If the length of the biomedical signal does not exceed 100,000 and the sample size is less than 200, the training process can be completed in no more five minutes.

The proposed PRS algorithm actually performed better than we had first expected. However, we are soberly aware that our study is not perfect and has a number of limitations. First, the rate of accuracy improvement is rather modest for each classification model. As shown in Fig. 6, a mere 1% improvement in accuracy is not enough to offset the error bar. One possible explanation for this might be a too-small sample size, resulting in inadequate model training [4]. Second, although the overall accuracy using the PRS features tends to be higher than that of the base and classical features sets for each classifier, there are still exceptions, as can be seen in Figs. 5–7. These phenomena are likely to be related to the possibility that extracting features based on the PRS algorithm does not work well in the presence of outliers. Root growth is based on the values in the soil matrix, which are obtained by performing statistical calculations on the biomedical signals. Mistakes are passed down, causing a reduction in accuracy. This would explain why, in some cases, lower accuracy is obtained at high training rates.

Although there remain many unanswered questions regarding the classification of biomedical signals, the PRS features proposed in this study, combined with proven classifiers, bring us closer to the cut-off point for clinical application. The limitations noted above will be addressed in subsequent studies to improve the algorithm while testing more biomedical signals.

## 6. Conclusion

Recognizing that existing feature extraction methods are not sufficient to produce biomedical signal classifiers that meet the requirements of clinical application, this research proposes a new swarm intelligence-based algorithm for extracting features to ensure higher classification accuracy. The proposed algorithm is based on the growth of roots in nature and offers good interpretability. Importantly, the features obtained by the PRS algorithm have a very low correlation with traditional features, and the accuracy of classifiers using features extracted by the proposed algorithm is shown to improve. This study lays the groundwork for future research focused on extending the use of biomedical signals from strictly research applications to clinical applications.